# Stochastic Domain Wall-Magnetic Tunnel Junction Artificial Neurons for Noise-Resilient Spiking Neural Networks

Thomas Leonard,[1,b)] Samuel Liu,[1,b)] Harrison Jin,[1] and Jean Anne C. Incorvia[1,a)]

[1]*Electrical and Computer Engineering, University of Texas at Austin, Austin, Texas 78712, USA*

The spatiotemporal nature of neuronal behavior in spiking neural networks (SNNs) make SNNs promising for edge applications that require high energy efficiency. To realize SNNs in hardware, spintronic neuron implementations can bring advantages of scalability and energy efficiency. Domain wall (DW) based magnetic tunnel junction (MTJ) devices are well suited for probabilistic neural networks given their intrinsic integrate-and-fire behavior with tunable stochasticity. Here, we present a scaled DW-MTJ neuron with voltage-dependent firing probability. The measured behavior was used to simulate a SNN that attains accuracy during learning compared to an equivalent, but more complicated, multi-weight (MW) DW-MTJ device. The validation accuracy during training was also shown to be comparable to an ideal leaky integrate and fire (LIF) device. However, during inference, the binary DW-MTJ neuron outperformed the other devices after gaussian noise was introduced to the Fashion-MNIST classification task. This work shows that DW-MTJ devices can be used to construct noise-resilient networks suitable for neuromorphic computing on the edge.

## INTRODUCTION

Neuromorphic computing allows for real time analysis of unstructured information that is too complex for conventional computing. Current digital implementations of neuromorphic computing rely on large numbers of CMOS transistors to simulate artificial neural networks (ANNs), which are inefficient in terms of both area and energy[1]. For edge computing, which can benefit greatly from energy efficient implementations, analog approaches to mapping synaptic weights and neuronal activations, *e.g.*, crossbar arrays of emerging memory devices, are promising solutions. Emerging devices can represent biological neurons and synapses on a one-to-one basis to exploit the computational dynamics found in the human brain[2–7]. Among them, spintronic devices offer a compact and energy efficient hardware-based building block for alternatives to digitally simulated neural networks on conventional CMOS devices[8].

Several types of magnetic devices have been proposed to emulate spiking neurons. These approaches are often focused on emulating the leaky integrate and fire (LIF) model of neuron behavior via gradual switching using Hall bars fabricated on the scale of tens of microns[9–12]. One such approach has incorporated the use of a domain wall (DW) spin texture, though with unfeasible area and latency[12]. Additionally, magnetic tunnel junctions are necessary for sufficiently large electrical readout signals. Scaled approaches for spintronic neurons have been proposed using alternative methods such as perpendicular magnetic tunnel junctions (MTJs)[13] and skyrmions[14–16], but have not been experimentally demonstrated. A promising approach is to

---

a) Author to whom correspondence should be addressed.  Electronic mail:  incorvia@austin.utexas.edu

b) Thomas Leonard and Samuel Liu contributed equally to this work.

integrate MTJ readout with scaled DW spin textures to represent the neuron state. These DW-MTJ devices offer a highly tunable way to emulate many neuromorphic functionalities on a monolithic platform. We have shown multi-weight (MW) artificial synapses with a single MTJ as well as directional synapses with tunable metaplasticity by varying the geometry of the DW track[2,17]. In addition, DW-MTJ devices are suited to emulate biological neurons given their intrinsic integrate-and-fire behavior, with the membrane potential represented by DW position. As current pulses are applied to the DW track, the DW can be moved using spin transfer torque (STT) or spin orbit torque (SOT). After a certain number of pulses, the DW will pass under the MTJ, causing a resistance change, which is able to represent a neuronal spike[18]. The ability of the DW-MTJ artificial neurons to incorporate other behaviors such as leaking[19,20], lateral inhibition[21,22], and edgy-relaxed neuronal behavior[23] have also been simulated. The potential of having MW synapses and LIF neurons on a monolithic platform can lead to hardware neural networks that incorporate advantages of magnetic materials for edge computing, namely radiation hardness, speed, and energy efficiency.

Here we propose a binary DW-MTJ artificial neuron that has stochastic behavior and is fabricated from the same material as previously demonstrated metaplastic MW synapses[2], opening the possibility for SNNs with fully spintronic matrix operations. The device, shown in Fig. 1, consists of a rectangular DW track with a free magnetic switching layer and a tantalum heavy metal layer beneath. The heavy metal layer allows for efficient SOT switching. At one end of the track lies an MTJ consisting of a thin MgO tunnel barrier and a synthetic antiferromagnet (SAF) pinned magnetic layer. The MTJ is positioned sufficiently far down the track to allow integration before firing. A bias field is used to assist the SOT switching to overcome the SAF pinning, but this could be removed via stack engineering. Furthermore, the device is reset using an external magnetic field after each fire event, but this could be removed by tuning the DW track geometry to facilitate intrinsic leaking. The material stack was grown using physical vapor deposition and patterned using electron beam lithography and ion beam etching, identically to Ref. 2. We show stable resistance states and repeatable cycle-to-cycle switching. Stochastic switching is demonstrated, and switching probability is shown to be dependent on the voltage pulse amplitude. The switching probability as a function of voltage amplitude is then used to form a lookup table to simulate the dynamics of the stochastic neurons on the network level using the Norse spiking neural network framework[24]. Our online learning results show that the binary neurons perform slightly worse than the LIF baseline and the MW neuron network. However, the difference between the MW device and the proposed binary neuron becomes negligible if the hidden layer size or the number of timestamps are reduced. This suggests that the binary stochastic neuron has application in edge computing, where space and energy consumption are highly constrained. This becomes clear when the same networks are used to perform inference. Using Fashion-MNIST with gaussian noise, we show that the binary neuron is more noise-resilient than both the MW device and the LIF baseline.

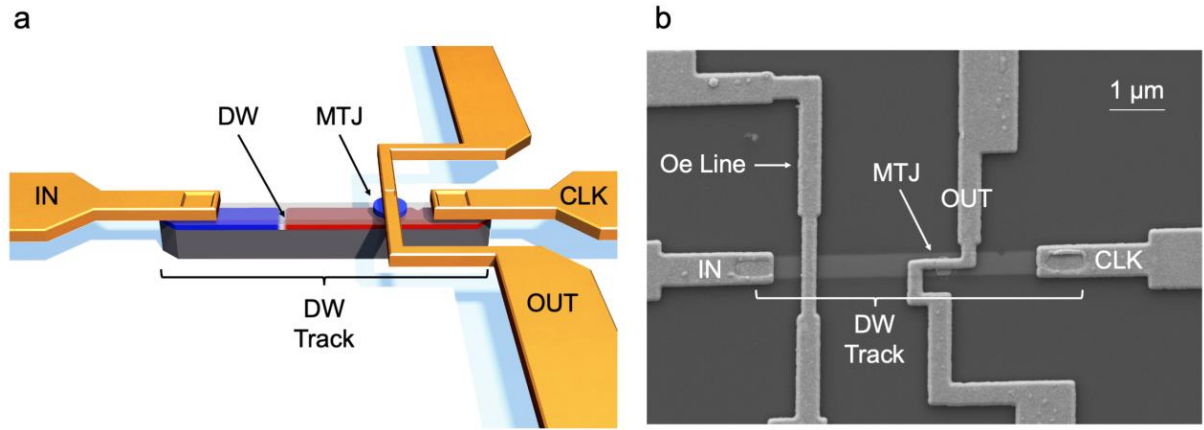

FIG. 1. Device schematic. (a) 3D rendering of the binary DW-MTJ stochastic neuron. Blue/red represent up/down magnetic domains with white DW. Grey represents Tantalum underlayer. The three critical terminals are IN, CLK, and OUT. (b) SEM image of the binary DW-MTJ stochastic neuron. The right notch is visible; the left notch is under the Oe line.

## EXPERIMENTAL DATA

Figure 1(a) shows a schematic of the studied binary DW-MTJ artificial neuron. In comparison to the logic devices in Ref. 18 the difference is the relative location of the MTJ along the DW track. This is shown in the scanning electron microscope (SEM) image Fig. 1(b). The material stack, detailed in Supplementary Fig. S1, and the fabrication process were the same as Refs. 2,18. The device operates using spin-orbit-torque (SOT) DW motion using the tantalum underlayer, and the resistance of the MTJ depends on the location of the DW within the track since the top layer of the MTJ is pinned by a synthetic antiferromagnetic layer. Two lithographically-patterned notches on either side of the MTJ stabilize the DW position in one of two places along the track. To operate the device, a voltage pulse is applied from CLK to IN to nucleate, depin, and drive the DW down the track. The resistance of the MTJ is then read from IN to OUT. The Oersted (Oe) line was not used during the testing since DW nucleation occurs probabilistically during writing and the Oe line reduces stochasticity.

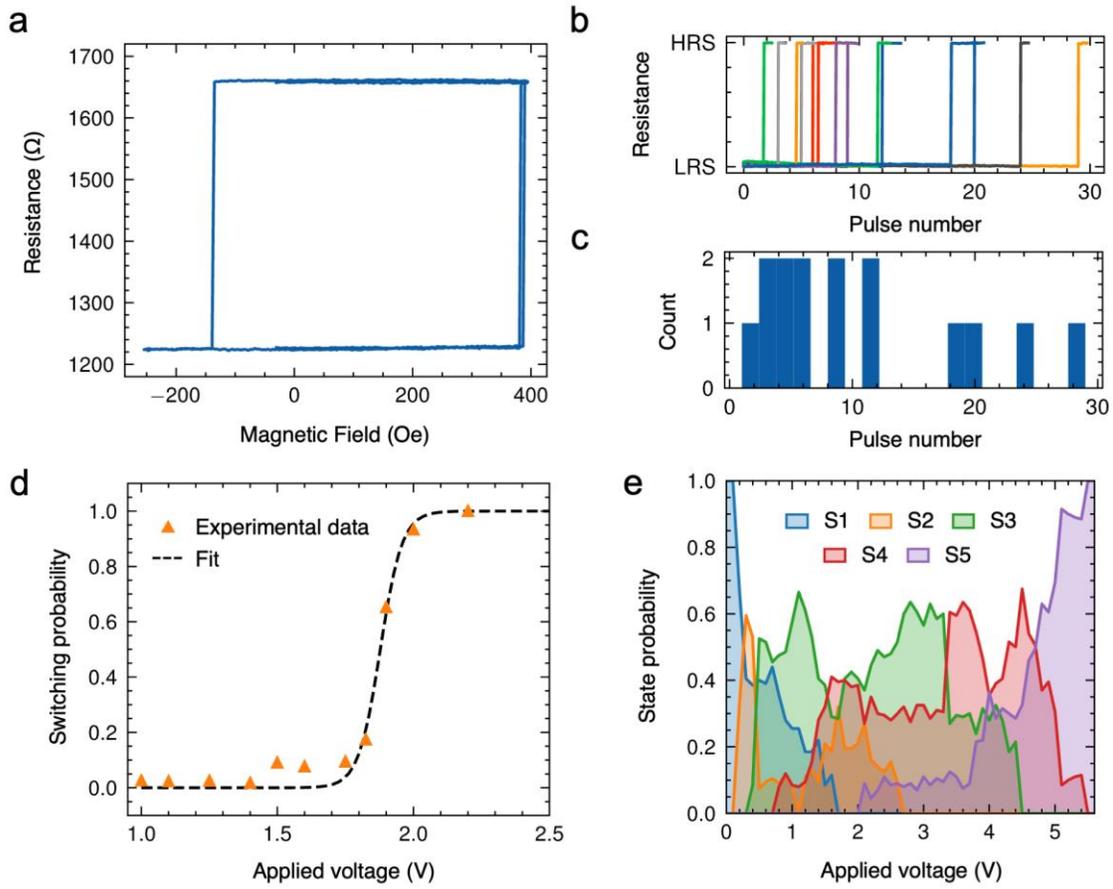

FIG. 2. Experimental results. (a) Binary neuron MTJ resistance (two-point resistance from IN to OUT) as function of out-of-plane magnetic field. (b) MTJ resistance vs. pulse number for constant pulse amplitude of 1.75 V repeatably applied from CLK to IN for 15 cycles. (c) Count of number of times each pulse number switches the binary DW-MTJ neuron, for 1.75 V constant amplitude. (d) Switching probability vs. applied voltage for respective (b-c) data obtained for 11 voltage amplitudes. (e) Analogous measured data for a MW DW-MTJ neuron from Ref. 2 which has 5 lithographic notches and hence a probability in being in one of five states, S1-S5, for a given applied volage pulse amplitude.

The binary DW-MTJ stochastic neuron data is shown in Fig. 2. The high resistance antiparallel state and the low resistance parallel state are shown in the minor field loop of Fig. 2(a). The device was set to the parallel state using an external magnetic field of -200 Oe. After saturation, a bias field of 350 Oe was used to aid in SOT switching. These field strengths were constant for all experiments. The bias field was not strong enough to switch the device by itself as shown in the field loop (Fig. 2(a)), but it reduced the voltage requirement to move the DW using SOT. This bias field could be removed through film stack optimization[25]. After the bias field was set, constant voltage pulses were applied from CLK to IN to nucleate, depin, and move the DW down the track. The pulse lengths were fixed at 30 ns in length with 10 ns rise/fall time. Figure 2(b) shows switching data for 15 cycles at a single voltage amplitude of 1.75 V. This cycling experiment was repeated for 11 different voltage amplitudes between 1.0 V to 2.2 V (Supplementary Fig. S2). After cycling the device, the data was tabulated as the number of times each pulse number switched the device for that voltage amplitude, shown in Fig. 2(c) for 1.75 V. The switching probability per pulse for each voltage amplitude is plotted in Fig. 2(d), showing a sigmoid fit.

The energy dissipation required for a write pulse can be estimated from this distribution. For a switching probability of around 50%, 1.8 V is applied across the heavy metal wire, which has a two-point resistance of 1029 Ω. The four-point resistance of the device is 365 Ω, indicating that 0.64 V is dissipated across the DW track. As a result, the estimated energy per sample is calculated to be around 44.9 pJ. This energy is competitive with digital SRAM implementations of spiking neurons at greatly reduced area and without standby energy dissipation from leakage current[26,27]. To further decrease the energy dissipation, shorter pulse length voltage pulses can be used. Additionally, the DW-MTJ neurons benefit from scaling, where the threshold voltage would decrease proportionally to a decrease in width of the DW track[28,29].

Figure 2(e) is data from our previous measurements of a multi-weight (MW) DW-MTJ device shown in Ref. 2 that is used as a comparison to the binary DW-MTJ measured in Fig. 2(a-d). Because the binary device presented here only has two states, integration is not emulated in device operation by the artificial neuron. The MW device was operated as a synapse in Ref. 2 (schematic and SEM image in Supplementary Fig. S3(b-c)), but the device can also be used as a neuron by reducing the size of the MTJ to cover only a small region just before the last notch, shown in Supplementary Fig. S3(d). The measured synapse device data is used to characterize the behavior of the hidden state of the simulated neuron, thereby emulating integration behavior. The MW DW-MTJ was tested using ramping voltages from 0 to 5.5 V for 10 cycles, and the voltage-dependent probability that the device is in any of the five states is tabulated. The results show that the MW DW-MTJ device combines quantized integration behavior with stochastic switching, showing the flexibility of the device platform.

**SPIKING NEURAL NETWORK SIMULATION**

In this section, we show how the binary stochastic DW-MTJ neuron can be used to build noise-resilient SNNs for training and inference on the edge. The architecture of the network is a multilayer perceptron (MLP) with two hidden layers with the stochastic neuron activations, shown in Fig. 3(a). The output layer is a leaky integration layer to allow for accurate softmax activations. To regulate the spiking frequency through each layer, a batch normalization layer is added after each hidden layer. The SNN is implemented in the Norse framework[24] with custom neuron modules using lookup tables consisting of the experimental data collected from the binary DW-MTJ (Fig. 2(d)) and MW DW-MTJ (Fig. 2(e)) stochastic neurons. To demonstrate supervised training performance, the spiking MLP was applied on the Fashion-MNIST clothing article classification task[30]. The Fashion-MNIST dataset was encoded in the frequency domain using a Poisson spiking encoding scheme, limiting the maximum spiking frequency of the brightest pixels to the sampling speed of the network, chosen to be 10 MHz. The loss function of the network was calculated by taking the maximum value across the timesteps of each output channel, then a softmax was applied to normalize. The Adam optimizer[31] was used to calculate the weight updates, using a learning rate of 0.001. The training was done with a batch size of 100 images.

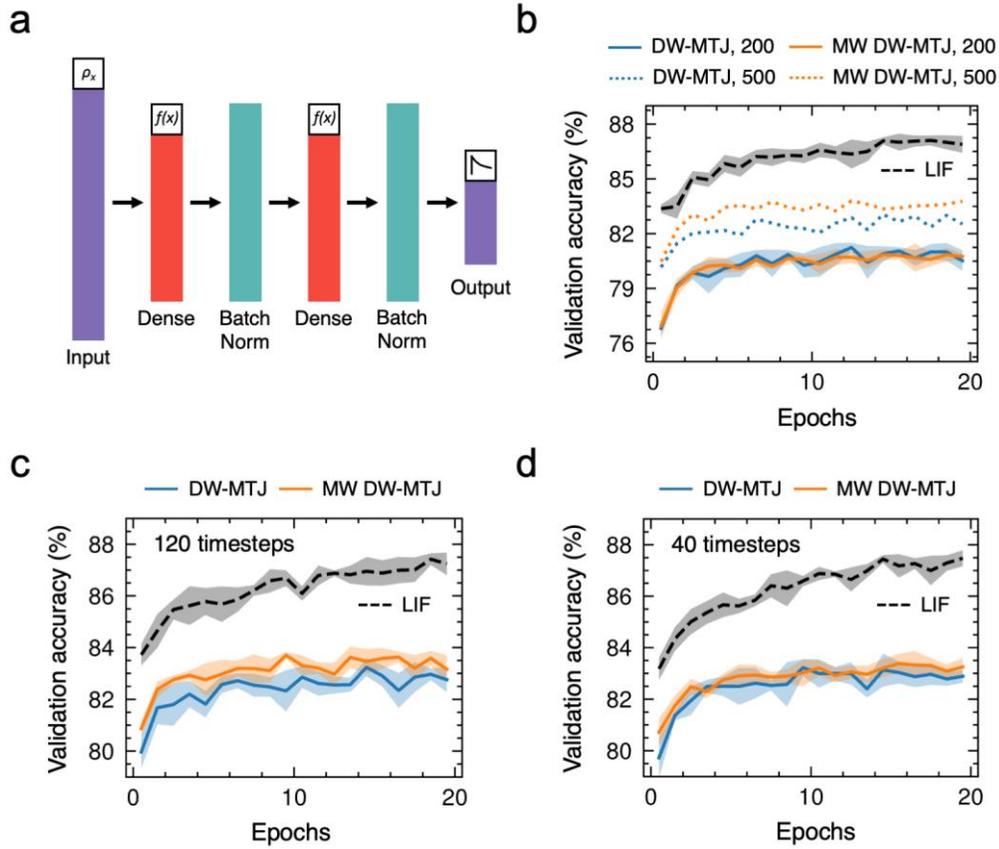

FIG. 3. Spiking neural network architecture and training performance on Fashion-MNIST. (a) Schematic of multilayer perceptron with 2 hidden layers. (b) Validation accuracy of networks consisting of binary DW-MTJ (blue) and MW DW-MTJ (orange) stochastic neurons for hidden layer sizes of 200 units (solid line) and 500 units (dotted line). LIF neuron SNN validation accuracy is also shown (dashed black line). Clouds depict the standard deviation of accuracy across 5 seeds. Validation accuracy of three network types with 500 hidden units trained for (c) 120 timesteps and (d) 40 timesteps per image shown.

Several networks with different activations were evaluated to compare the training performance of the binary DW-MTJ and MW DW-MTJ stochastic neurons, compared to a baseline of an ideal LIF neuron. Figure 3b shows SNNs consisting of the three types of neurons with 200 (dotted lines) and 500 units (solid lines) per hidden layer. The network was sampled for 80 timesteps. The training performance results for the baseline LIF network are of a network with 500 units per hidden layer, though reducing the number of units per hidden layer to 200 did not significantly affect the validation accuracy. Because the MW DW-MTJ had multiple levels, a form of quantized integrate-and-fire behavior was emulated. This extra expressivity added complexity to the activation function, allowing the MW DW-MTJ neuron to slightly outperform the simpler binary DW-MTJ neuron when the network had 500 units per hidden layer. However, in the network with 200 units per hidden layer, there was no significant difference in training performance between the two networks, suggesting that the expressivity of the extra levels in the MW DW-MTJ device does not significantly improve training performance for smaller networks. This indicates that the simpler fabrication of the binary DW-MTJ neuron can still be desirable for resource-limited edge computing.

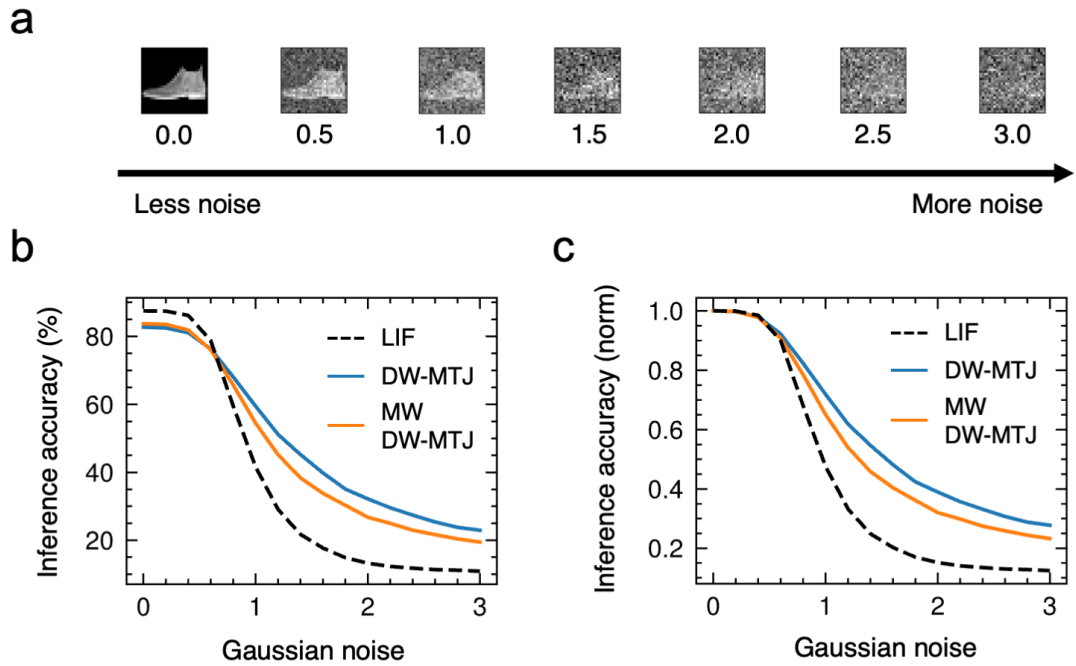

FIG. 4. Inference accuracy on Fashion-MNIST data with (a) added gaussian noise. (b) Raw and (c) normalized test accuracy for added gaussian noise with standard deviation from 0 to 3.

This is further corroborated by the validation accuracy results where networks with 500 hidden units per layer were sampled for 120 (Fig. 3(c)) and 40 (Fig. 3(d)) timesteps. In the network sampled for 120 timesteps, the MW DW-MTJ device network slightly outperformed the binary DW-MTJ in terms of validation accuracy. However, when the network was sampled for 40 timesteps, both types of DW-MTJ devices had similar validation accuracy. As a result, for smaller networks sampled with less timesteps, it can be beneficial to reduce fabrication complexity by choosing the binary DW-MTJ instead of the MW device.

It is well known in literature that when noise is introduced during the training phase, the resulting network is less affected by noisy input data[32]. This is because the optimization algorithm will find a network state that will minimize the error in a noise-resilient way during training. Due to the probabilistic nature of switching, both types of DW-MTJ neurons are hypothesized to be able to fulfill the same functionality of increasing the noise-resilience of a network. To characterize this, varying degrees of gaussian noise were added to the 10,000 images in the Fashion-MNIST test set, shown in Fig. 4(a). Inference is then performed on the noisy dataset using the highest accuracy models of each type trained on the non-noisy dataset in the previous section. The results in Fig. 4(b) show that while the LIF-based network maintained a higher starting accuracy than the other two networks, there was a sharp drop off at a gaussian noise magnitude of 0.7 to below that of the stochastic DW-MTJ neurons. This point is also where the inference accuracy of the network composed of MW DW-MTJ devices dropped below that of the simpler binary DW-MTJ devices. At a gaussian noise magnitude of 3.0, the LIF network accuracy became insignificant compared to a random guess (~10%) while both DW-MTJ device-based networks are still able to accomplish inference at a better accuracy than a random guess. The advantage of the DW-MTJ stochastic neurons becomes clearer when viewing the data normalized to the accuracy of the networks on pristine data (Fig. 4(c)). In practical implementation, this noise resilience may not be an acceptable tradeoff for accuracy. However, in deep neural networks for edge applications, a promising implementation is to integrate stochastic neurons in one layer to introduce noise-resilience while maintaining high classification accuracy through LIF neurons in other layers.

## CONCLUSION

In summary, scaled DW-MTJ artificial neurons were fabricated for the first time, and stochastic switching behavior was characterized. The resulting distribution of spiking probability as a function of input voltage for the binary DW-MTJ device was approximately sigmoidal, with a low energy dissipation of 44.9 pJ per sample for an individual device. An alternative binarized device architecture was also proposed, with multiple notches to implement quantized integration states. The experimental data for both devices was then mapped into lookup tables to simulate supervised learning of a MLP for SNNs. Online learning results show that while there was a slight (~1%) advantage in validation accuracy to training with the MW DW-MTJ due to the additional expressivity enabled by the extra notches. However, at smaller network sizes and fewer timesteps the binary DW-MTJ device can match the performance of the MW DW-MTJ device. When performing inference on noisy datasets, the DW-MTJ devices were able to outperform the ideal LIF neurons due to the introduced noise resilience in networks that were trained using the stochastic devices. Where the LIF neuron network inference accuracy rapidly decayed with increasing noise in the dataset, the networks composed of the DW-MTJ neurons of both types were able to maintain reasonable performance above that of the LIF network. These results indicate that the proposed devices are well suited for energy efficient, resource-limited hardware for neuromorphic computing at the edge, making a further case for analog DNN accelerators with spintronic weights and activations.

## ACKNOWLEDGMENTS

The authors acknowledge research support from the National Science Foundation under CCF award 1910997, as well as support from the National Science Foundation graduate research fellowship program under Grant Nos. 2020307514 and 2021311125. The work was done at the Texas Nanofabrication Facility supported by NSF Grant No. NNCI- 1542159 and at the Texas Materials Institute (TMI). The authors acknowledge computing resources from the Texas Advanced Computing Center (TACC) at the University of Texas at Austin (http://www.tacc.utexas.edu).

## DATA AVAILABILITY

The data that support the findings of this study are available from the corresponding author upon reasonable request.